\documentclass[10pt,twocolumn,letterpaper]{article}

\usepackage[pagenumbers]{cvpr} 

\usepackage{graphicx}
\usepackage{amsmath}
\usepackage{amssymb}
\usepackage{booktabs}
\usepackage{bbm}
\usepackage[dvipsnames]{xcolor}
\usepackage{arydshln}
\usepackage{multirow}
\usepackage{footmisc}
\usepackage{algorithm}
\usepackage{algpseudocode}

\usepackage{pifont}
\newcommand{\cmark}{\ding{51}}%
\newcommand{\xmark}{\ding{55}}%

\usepackage{hyperref}
\hypersetup{colorlinks,breaklinks}

\usepackage[capitalize]{cleveref}
\crefname{section}{Sec.}{Secs.}
\Crefname{section}{Section}{Sections}
\Crefname{table}{Table}{Tables}
\crefname{table}{Tab.}{Tabs.}

\def\cvprPaperID{} 
\def\confName{CVPR}
\def\confYear{2023}

\begin{document}

\title{GFPose: Learning 3D Human Pose Prior with Gradient Fields}

\author{Hai Ci$^1$, \quad Mingdong Wu$^{2,1}$, \quad 
Wentao Zhu$^{2}$, \quad Xiaoxuan Ma$^{2}$, \\
Hao Dong$^{2}$, \quad Fangwei Zhong$^{3,1}$, 
\quad Yizhou Wang$^{2,1}$\\
\textsuperscript{1} Beijing Institute for General Artificial Intelligence\\
\textsuperscript{2} CFCS, School of Computer Science, Peking University\\
\textsuperscript{3} School of Intelligence Science and Technology, Peking University\\
{\tt\small cihai@bigai.ai, \{wmingd, wtzhu, maxiaoxuan, hao.dong, zfw, yizhou.wang\}@pku.edu.cn}
}
\maketitle
\newcommand\mingdong[1]{\textcolor{teal}{WMD: #1}}

\newcommand{\loss}{\mathcal{L}}
\newcommand{\E}{\mathbb{E}}
\newcommand{\R}{\mathbb{R}}
\newcommand{\x}{x}
\newcommand{\z}{\mathbf{z}}
\newcommand{\N}{\mathcal{N}}
\newcommand{\X}{\mathcal{X}}
\newcommand{\C}{\mathcal{C}}
\newcommand{\data}{p_{data}}
\newcommand{\cond}{\mathbf{c}}
\newcommand{\pose}{\mathbf{x}}
\newcommand{\score}{\mathbf{s}_{\theta}}
\newcommand{\pscore}{\mathbf{\Phi}_{\theta}}
\newcommand\norm[1]{\left\lVert#1\right\rVert}

\newcommand{\npose}{\mathbf{\widetilde \x}}

\def\eg{\emph{e.g}.} \def\Eg{\emph{E.g}.}
\def\ie{\emph{i.e}.} \def\Ie{\emph{I.e}.}
\def\cf{\emph{c.f}.} \def\Cf{\emph{C.f}.}
\def\etc{\emph{etc}.} \def\vs{\emph{vs}.}
\def\wrt{w.r.t. } \def\dof{d.o.f. }
\def\etal{\emph{et al}. }

\begin{abstract}
    Learning 3D human pose prior is essential to human-centered AI.
    Here, we present GFPose, a versatile framework to model plausible 3D human poses for various applications.
    At the core of GFPose is a time-dependent score network, which estimates the gradient on each body joint and progressively denoises the perturbed 3D human pose to match a given task specification. 
    During the denoising process, GFPose implicitly incorporates pose priors in gradients and unifies various discriminative and generative tasks in an elegant framework. 
    Despite the simplicity, GFPose demonstrates great potential in several downstream tasks. 
    Our experiments empirically show that 1) as a multi-hypothesis pose estimator, GFPose outperforms existing SOTAs by 20\% on Human3.6M dataset. 2) as a single-hypothesis pose estimator, GFPose achieves comparable results to deterministic SOTAs, even with a vanilla backbone. 3) GFPose is able to produce diverse and realistic samples in pose denoising, completion and generation tasks.\footnote{Project page \url{https://sites.google.com/view/gfpose/}}

    
\end{abstract}

\vspace{-1.0em}
\section{Introduction}
\label{sec:introduction}
Modeling 3D human pose is a fundamental problem in human-centered applications, \eg \ augmented reality~\cite{lin2010augmented, murphy2010head}, virtual reality~\cite{ahuja2019mecap, yang2022hybridtrak, mehta2017vnect}, and human-robot collaboration~\cite{cheng2021human, gao2019dual, liu2021collision}. Considering the biomechanical constraints, natural human postures lie on a low-dimensional manifold of the physical space. Learning a good prior distribution over the valid human poses not only helps to discriminate the infeasible ones but also enables sampling of rich and diverse human poses. The learned prior has a wide spectrum of use cases with regard to recovering the 3D human pose under different conditions, \eg, monocular images with depth ambiguities and occlusions~\cite{cheng2019occlusion, wehrbein2021probabilistic}, inertial measurement unit (IMU) signals with noises~\cite{zhang2020fusing}, or even partial sensor inputs~\cite{winkler2022questsim, jiang2022avatarposer}.





\begin{figure}[!t]
    \centering
    \includegraphics[width=0.97\linewidth]{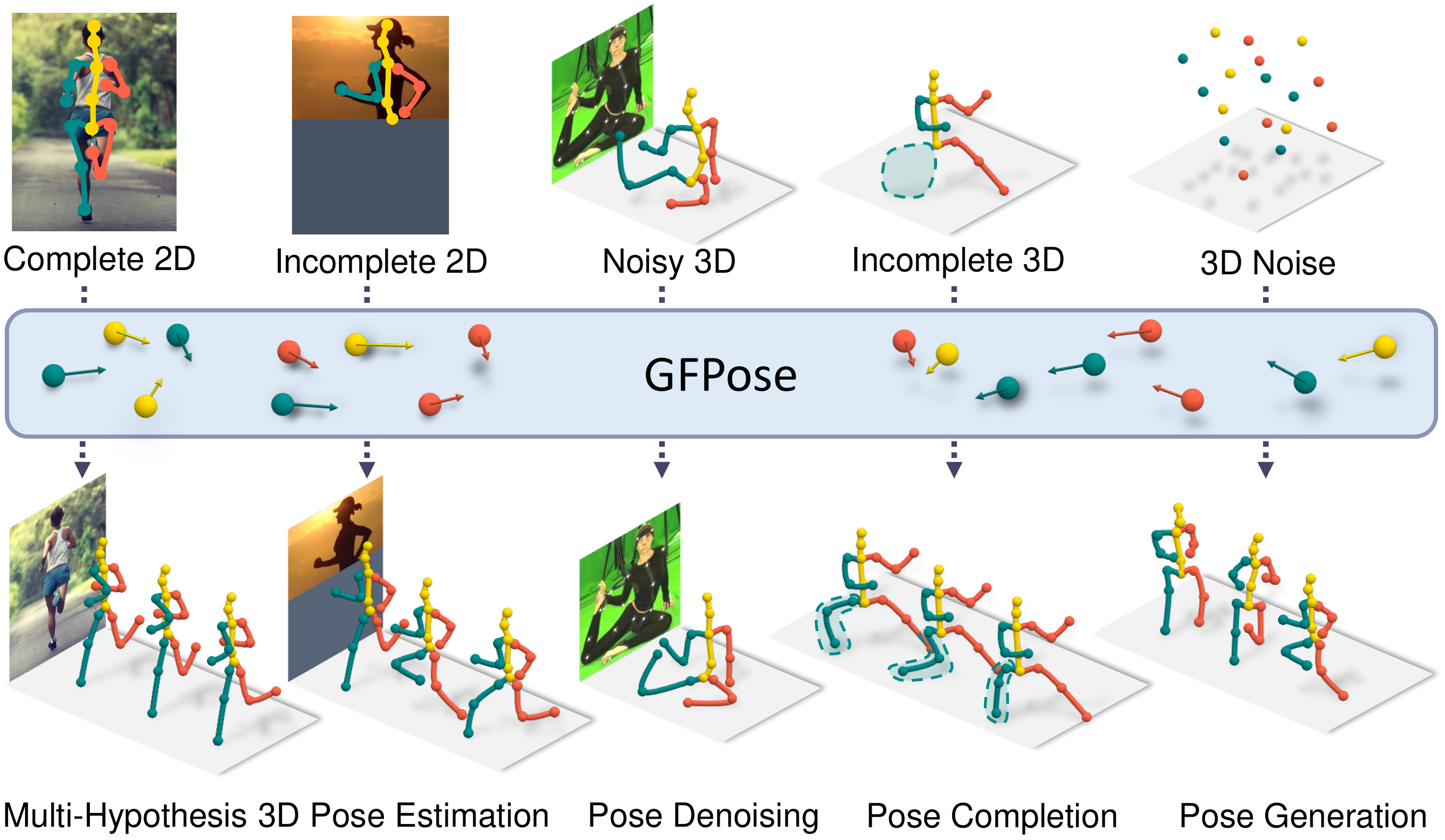}
        \caption{GFPose learns the 3D human pose prior from 3D human pose datasets and represents it as gradient fields for various applications, \eg, multi-hypothesis 3D pose estimation from 2D keypoints, correcting noisy poses, completing missing joints, and generating natural poses from noise.}
    \label{fig:teaser}
    \vspace{-1.0em}
\end{figure}

Previous works explore different ways to model human pose priors. Pioneers~\cite{hatze1997three, kodek2002identifying} attempt to explicitly build joint-angle limits based on biomechanics. Unfortunately, the complete configuration of pose-dependent joint-angle constraints for the full body is unknown. 
With the recent advances in machine learning, a rising line of works seek to learn the human pose priors from data. Representative methods include modeling the distribution of plausible poses with GMM\cite{bogo2016keep}, VAE\cite{pavlakos2019expressive}, GAN\cite{davydov2022adversarial} or neural implicit functions\cite{tiwari2022pose}.
These methods learn an independent probabilistic model or energy function to characterize the data distribution $\data(\pose)$. They usually require additional optimization process to introduce specific task constraints when applied to downstream tasks. Therefore extra efforts such as balancing prior terms and different task objectives are inevitable. Some methods jointly learn the pose priors and downstream tasks via adversarial training~\cite{hmrKanazawa17, kocabas2019vibe} or explicit task conditions~\cite{ling2020character, rempe2021humor, petrovich21actor} $\data(\pose | \cond)$. These methods seamlessly integrate priors into learning-based frameworks, but limit their use to a single given task.

In this work, we take a new perspective to learn a versatile 3D human pose prior model for general purposes. 
Different from previous works that directly model the plausible pose distribution $p_{data}(\pose)$, we learn the score (gradient of a log-likelihood) of a task conditional distribution $\nabla_{\pose}\log\data(\pose | \cond )$, where $\cond$ is the task-specific condition, \eg, for 3D human pose estimation, $\cond$ could be 2D images or detected 2D poses. $\pose$ represents plausible 3D human poses. 
In this way, we can jointly encode the human pose prior and the task specification into the score, instead of considering the learned prior model as an ad-hoc plugin as in an optimization process. 
To further enhance the flexibility and versatility, we introduce a condition masking strategy, where task conditions are randomly masked to varying degrees during training. Different masks correspond to different task specifications. Thus we can handle various pose-related tasks in a unified learning-based framework.

We present GFPose, a general framework for pose-related tasks. GFPose learns a time-dependent score network $\score(\pose, t|\cond)$ to approximate $\nabla_{\pose}\log\data(\pose | \cond )$ on a large scale 3D human pose dataset~\cite{h36m_pami} via Denoising Score Matching~(DSM) \cite{denosingScoreMatching, hyvarinen2005estimation, song2020sliced, song2019generative, song2020improved, song2020score, song2021maximum}. 
Specifically, for any valid human pose $\pose \in \R^{J \times 3}$ in Euclidean space, we sample a time-dependent noise $\z(t)$ from a prior distribution, perturb $\pose$ to get the noisy pose $\npose$, then train $\score(\npose, t|\cond)$ to learn the score towards the valid pose. Intuitively, the score points in the direction of increasing pose plausibility. 
To handle a wider range of downstream tasks, we adopt a hierarchical condition masking strategy in training. Concretely, we randomly mask out the task condition $\cond$ by sampling masks from a hierarchy of candidate masks. The candidate masks cover different levels of randomness, including human level, body part level, and joint level. This helps the model to build the spatial relation between different body joints and parts, and enables GFPose directly applicable to different task settings at test time (Figure~\ref{fig:teaser}), \eg, recovering 3D pose from severe occlusions when $\cond$ is partially masked 2D pose or unconditional pose generation when $\cond$ is fully masked ($\cond = \O $).

We evaluate GFPose on various downstream tasks, including monocular 3D human pose estimation, pose denoising, completion, and generation. Empirical results on the H3.6M benchmark~\cite{h36m_pami} show that: 
1) GFPose outperforms SOTA in both multi-hypothesis and single-hypothesis pose estimation tasks~\cite{wehrbein2021probabilistic} and demonstrates stronger robustness to severe occlusions in pose completion~\cite{Li_2019_CVPR}. Notably, under the single-hypothesis setting, GFPose can achieve comparable pose estimation performance to previous deterministic SOTA methods~\cite{zeng2020srnet, ci2019optimizing, pavllo20193d} that learns one-to-one mapping. To the best of our knowledge, this is for the first time that a probabilistic model can achieve such performance. 2) As a pose generator, GFPose can produce diverse and realistic samples that can be used to augment existing datasets.
We summarize our contributions as follows:
\begin{itemize}
    \setlength\itemsep{-0.25em}
    \item We introduce GFPose, a novel score-based generative framework to model plausible 3D human poses.
    \item We design a hierarchical condition masking strategy to enhance the versatility of GFPose and make it directly applicable to various downstream tasks.
    \item We demonstrate that GFPose outperforms SOTA on multiple tasks under a simple unified framework.
\end{itemize}

\begin{figure*}[!t]
    \centering
    \includegraphics[width=0.97\linewidth]{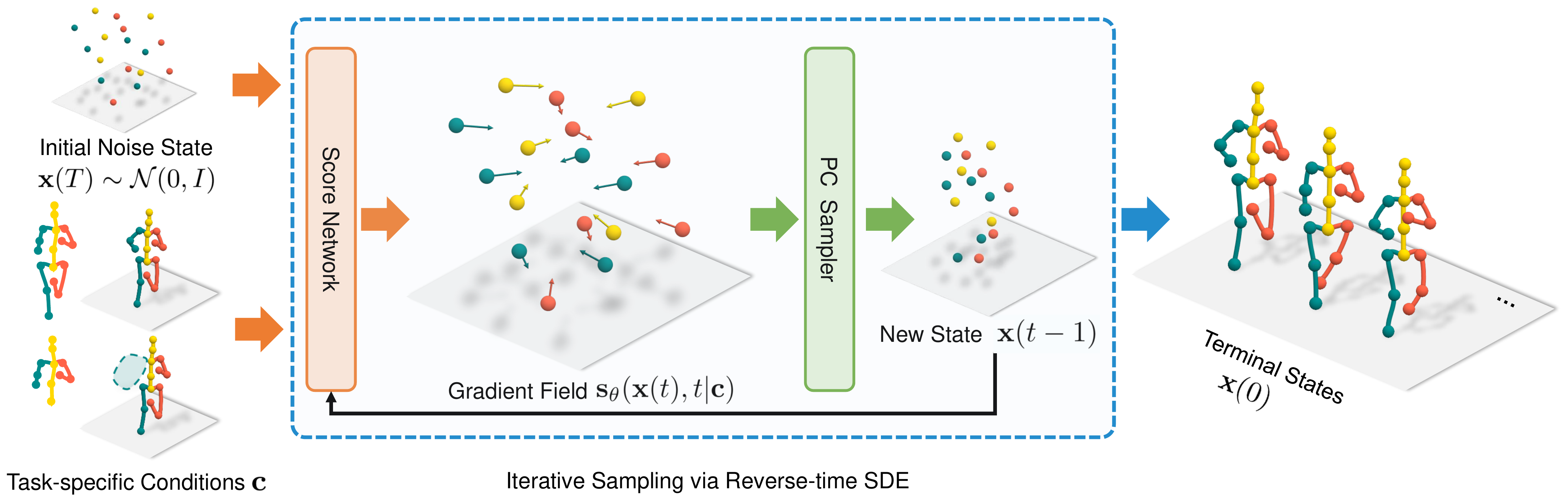}
        \caption{Inference pipeline of GFPose. For a downstream task specified by condition $ \cond$, we generate terminal states $\pose(0)$ from initial noise states $\pose(T)\sim \N(0, I)$ via a reverse-time Stochastic Differential Equation (RSDE). $T$ denotes the start time of RSDE. This process is simulated by a Predictor-Corrector sampler\cite{song2020score}. At each time step $t$, the time-dependent score network $\score(\pose(t), t | \cond)$ outputs gradient fields that `pull' the pose to be more valid and faithful to the task condition $ \cond$.}
    \label{fig:pipeline}
    \vspace{-1.25em}
\end{figure*}

\vspace{-1.0em}
\section{Related Work}
\label{sec:related_work}

\subsection{Human Pose Priors}

We roughly group previous works on learning 3D human pose priors into two categories. The first line of works learns task-independent pose priors, \ie they learn the unconditional distribution $\data(\pose)$ of plausible poses. These approaches usually involve time-consuming optimization process to introduce task-specific constraints when applying the learned priors to different downstream tasks. Akheter \etal~\cite{akhter2015pose} learns the pose-dependent joint-angle limits directly. SMPLify~\cite{bogo2016keep} fits a mixture of Gaussians to motion capture (mocap) data. VAE-based methods such as VPoser~\cite{pavlakos2019expressive} map the 3D poses to a compact representation space and can be used to generate valid poses. GAN-based methods\cite{davydov2022adversarial} learn adversarial priors by discriminating the generated poses from the real poses. Most recently, Pose-NDF~\cite{tiwari2022pose} proposes to model the plausible pose manifolds with a neural implicit function. 
The second line of works focuses on task-aware priors. They learn the conditional priors $\data(\pose | \cond)$ under specific task constraints. MVAE\cite{ling2020character} and HuMoR~\cite{rempe2021humor} employ autoregressive conditional VAE to learn plausible pose trajectories given the historical state. ACTOR~\cite{petrovich21actor} learns an action-conditioned variational prior using Transformer VAE.
HMR~\cite{hmrKanazawa17} and VIBE~\cite{kocabas2019vibe} jointly train the model with the adversarial prior loss and pose reconstruction loss.
In this paper, we explore a novel score-based framework to model plausible human poses. Our method jointly learns the task-independent and various task-aware priors via a hierarchical condition masking strategy. Thus it is native and directly applicable to multiple downstream tasks without involving extra optimization steps.

\subsection{3D Human Pose Estimation}
Estimating the 3D human pose from a monocular RGB image is a fundamental yet unresolved problem in computer vision. The common two-stage practice is estimating the 2D pose with an off-the-shelf estimator first, then lifting it to 3D, which improves the model generalization. Most existing methods directly learn the 2D-to-3D mapping via various architecture designs~\cite{martinez_2017_3dbaseline, lcn-pami, xu2021graph, zhaoCVPR19semantic, ma2021context}. Nonetheless, solving the 2D-to-3D mapping is intrinsically an ill-posed problem as infinite solutions suffice without extra constraints. Therefore, formulating and learning 3D human pose estimation as a one-to-one mapping fails to express the ambiguity and inevitably suffers from degraded precision~\cite{DBLP:conf/eccv/DongSZLZB20, duanrevisiting}. To this end, Li~\etal\cite{Li_2019_CVPR} propose a multimodal mixture density network to learn the plausible pose distribution instead of the one-to-one mapping. Jahangiri~\etal\cite{jahangiri2017generating} design a compositional generative model to generate multiple 3D hypotheses from 2D joint detections. Sharma~\etal \cite{Sharma_2019_ICCV} first sample plausible candidates with a conditional variational autoencoder, then use ordinal relations to filter and fuse the candidates. 
\cite{kolotouros2021prohmr,WehRud2021,pierzchlewicz2022multi} employs normalizing flows to model the distribution of plausible human poses.
Li~\etal\cite{li2022mhformer} use a transformer to learn the spatial-temporal representation of multiple hypotheses.
We show that GFPose is a suitable solution for multi-hypothesis 3D human pose estimation, and is able to handle severe occlusions by producing plausible hypotheses.




\subsection{Score-Based Generative Model}
The score-based generative model aims at estimating the gradient of the log-likelihood of a given data distribution~\cite{denosingScoreMatching, hyvarinen2005estimation, song2020sliced, song2019generative, song2020improved, song2020score, song2021maximum}. 
To improve the scalability of the score-based generative model, \cite{song2020sliced} introduces a sliced score-matching objective that projects the scores onto random vectors before comparing them.
Song \etal introduce annealed training for denoising score matching~\cite{song2019generative} and several improved training techniques~\cite{song2020improved}. They further extend the discrete levels of annealed score matching to a continuous diffusion process and show promising results on image generation~\cite{song2020score}.
These recent advances promote the wide application of the score-based generative model in different fields, such as object rearrangement~\cite{wu2022targf}, medical imaging~\cite{song2021solving}, point cloud generation~\cite{cai2020learning}, molecular conformation generation~\cite{shi2021learning}, scene graph generation~\cite{suhail2021energy}, point cloud denoising~\cite{luo2021score}, offline reinforcement learning~\cite{janner2022planning}, and depth completion~\cite{shao2022diffustereo}.
Inspired by these promising results, we seek to develop a score-based framework to model plausible 3D human poses.
To the best of our knowledge, our method is the first to explore score-based generative models for learning 3D human pose priors.

\section{Revisiting Denoising Score Matching}
\label{sec:preliminary}
Given samples $\{ \pose_i \}_{i=1}^N$ from an unknown data distribution $\{ \pose_i \sim p_{data}(\pose) \}$, 
the score-based generative model aims at learning a \textit{score function} to approximate $\nabla_{\pose} \log p_{data}(\pose)$ via a \textit{score network} $\score(\pose): \R^{|\X|} \rightarrow \R^{|\X|}$.
\begin{equation}
    \loss(\theta) = \frac{1}{2}\E_{p_{data}}\left[||\score(\pose) - \nabla_{\pose} \log p_{data}(\pose)||^2_2\right].
\label{eq: Vanilla Score-matching}
\end{equation}
During the test phase, a new sample is generated by Markov chain Monte Carlo (MCMC) sampling, \eg, Langevin Dynamics (LD). Given a step size $\epsilon > 0$, an initial point $\tilde\pose_0$ and a Gaussian noise $\z_t \sim \N(0, I)$, LD can be written as:
\begin{equation}
    \tilde\pose_t = \tilde\pose_{t-1} + \frac{\epsilon}{2}\nabla_{\pose}\log p_{data}(\tilde\pose_{t-1}) + \sqrt{\epsilon} \z_t.
\label{eq: Langevin Dynamics}
\end{equation}
When $\epsilon \rightarrow 0$ and $t \rightarrow \infty$, the $\tilde\pose_t$ becomes an exact sample from $p_{data}(\pose)$ under some regularity conditions~\cite{welling2011bayesian}.

However, the vanilla objective of score-matching in Eq.~\ref{eq: Vanilla Score-matching} is intractable, since $p_{data}(\pose)$ is unknown. To this end, the Denoising Score-Matching (DSM)~\cite{denosingScoreMatching} proposes a tractable objective by pre-specifying a noise distribution $q_{\sigma}(\widetilde \pose|\pose)$, \eg, $\N(0, \sigma^2I)$, and train a score network to denoise the perturbed data samples:
\begin{equation}
    \loss(\theta) = \E_{\widetilde \pose \sim q_{\sigma}, \pose \sim p_{data}}\left[||\score(\widetilde \pose) - \nabla_{\widetilde \pose}\log q_{\sigma}(\widetilde \pose|\pose) ||^2_2\right]
\label{eq: DSM-Gaussian}
\end{equation}
where $\nabla_{\widetilde \pose}\log q_{\sigma}(\widetilde \pose|\pose) = \frac{1}{\sigma^2}(\pose - \widetilde \pose)$ are tractable for the Gaussian kernel. DSM guarantees that the optimal score network holds $\score^*(\pose) = \nabla_{\pose} \log p_{data}(\pose)$ for almost all $\pose$.

\section{Method}
\label{sec:method}
\subsection{Problem Statement}
\label{sec:problem statement}

We seek to model 3D human pose priors under different task conditions $\data(\pose| \cond)$ by estimating $\nabla_{\pose}\log\data(\pose | \cond )$ from a paired dataset $\{ (\pose, \cond)\}^N$, where $\pose \in \R^{J \times 3}$ represents plausible 3D human poses and $\cond$ denotes different task conditions. In this work, we consider $\cond$ to be 2D poses ($\cond \in \R^{J \times 2}$) for monocular 3D human pose estimation tasks; 3D poses ($\cond \in \R^{J \times 3}$) for 3D pose completion; $\O$ for pose generation and denoising. (elaborate in Section \ref{sec:experiments}) We further introduce a condition masking strategy to unify different task conditions and empower the model to handle occlusions.
Notably, this formulation does not limit to the choices used in this paper. In general, $\cond$ can be any form of observation, \eg, image features or human silhouettes for recovering 3D poses from the image domain. $\pose$ could also be different forms of 3D representation, \eg, joint rotation in SMPL~\cite{loper2015smpl}. With a learned prior model, different downstream tasks can be formulated as a unified generative problem, \ie, generate new samples from $\data(\pose| \cond)$.

\subsection{Learning Pose Prior with Gradient Fields}
\label{sec:learn_gf}

We adopt an extension~\cite{song2020score} of Denoising Score Matching (DSM)~\cite{denosingScoreMatching} to learn the score $\nabla_{\pose}\log\data(\pose | \cond )$ and sample plausible poses from the data distribution $\data(\pose| \cond)$. 
The whole framework consists of a forward diffusion process and a reverse sampling process:
(1) The forward diffusion process perturbs the 3D human poses from the data distribution to a predefined prior distribution, \eg, Gaussian distribution.
(2) The reverse process samples from the prior distribution and reverse the diffusion process to get a plausible pose from the data distribution. 

\noindent
\textbf{Perturb Poses via SDE}
Following \cite{song2020score}, we construct a time-dependent diffusion process $\{ \pose (t) \}^{T}_{t=0}$ indexed by a continuous time variable $t \in [0, T]$. $\pose(0) \sim p_0$ comes from the data distribution.
 $\pose(T) \sim p_T$ comes from the diffused prior distribution. As $t$ grows from $0$ to $T$, we gradually perturb the poses with growing levels of noise. The perturbation procedure traces a continuous-time stochastic process and can be modeled by the solution to an It$\hat{o}$ SDE:
\begin{equation}
    d\pose = \mathbf{f}(\pose, t) dt + g(t) d\mathbf{w}
\end{equation}
where $\mathbf{f}(\cdot, t): \R^d \to \R^d $ is called the \textit{drift coefficient}, $g(t) \in \R$ is called the \textit{diffusion coefficient} of $\pose(t)$. $dt$ represents infinitesimal time step. $\mathbf{w}$ is the Brownian motion, and $d\mathbf{w}$ can be seen as infinitesimal white noise. We have various designs of SDEs to perturb the pose $\pose$, \ie, different choices of $\mathbf{f}(\cdot, t)$ and $g(t)$. In this work, we use the subVP SDE 
\footnote{\label{footnote1}We detail subVPSDE, derivation of objective function and sampling process in the Supplementary.} 
proposed in \cite{song2020score}, which perturbs any human poses $\pose(0)$ to a Gaussian distribution $p_T$. 

\noindent
\textbf{Sample Poses via Reverse-Time SDE}
If we reverse the perturbation process, we can get a pose sample $\pose(0) \sim p_0$ from a Gaussian noise $\pose(T) \sim p_T$. According to \cite{anderson1982reverse,song2020score}, the reverse is another diffusion process described by the reverse-time SDE (RSDE):
\begin{equation}
\begin{aligned}
    d\pose = [\mathbf{f}(\pose, t) - g^2(t)\nabla_{\pose}\log p_{t}(\pose | \cond )] dt + g(t) d\bar{\mathbf{w}}
\label{eq:reverse_sde}
\end{aligned}
\end{equation}
where $t$ starts from $T$ and flows back to $0$. $dt$ here represents negative time step and $\bar{\mathbf{w}}$ denotes Brownian motion at reverse time. In order to simulate Eq. \ref{eq:reverse_sde}, we need to know $\nabla_{\pose}\log p_{t}(\pose | \cond )$ for all $t$. We train a neural network to estimate it.

\noindent
\textbf{Train Score Estimation Network}
According to Eq. \ref{eq: DSM-Gaussian}, we train a time-dependent score network $\score(\pose, t | \cond): \X \times \R^+ \times \C \rightarrow \X $ to estimate $\nabla_{\pose}\log p_{t}(\pose | \cond )$ for all $t$, where $\C$ denotes the condition space. The objective can be written as:
\begin{equation}
\begin{aligned}
   \E_{t\sim \mathcal{U}(0, T)}
   \{
   \lambda(t)\E_{\pose(0) \sim p_0(\pose|\cond), \pose(t) \sim p_{0t}(\pose(t) | \pose(0) , \cond)}
   \\ 
   [\|\score(\pose(t), t | \cond)  - \nabla_{\pose(t)}\log p_{0t}(\pose(t) \mid \pose(0) , \cond)\|_2^2 ] \}
\end{aligned}
\label{eq: SDE score matching}
\end{equation}
where $t$ is uniformly sampled over $[0, T]$. $\lambda(t)$ is a weighting term. $p_{0t}$ denotes the perturbation kernel. Due to the choice of subVP SDE, we can get a closed form of $p_{0t}$\footref{footnote1}. Thus, we can get a tractable objective of Eq.~\ref{eq: SDE score matching}. 

Given a well-trained score network $\score(\pose, t | \cond)$, we can iterate over Eq.~\ref{eq:reverse_sde} to sample poses from $\data(\pose|\cond)$ as illustrated in Fig. \ref{fig:pipeline}. At each time step $t$, the network takes the current pose $\pose(t)$, time step $t$ and task condition $\cond$ as input and outputs the gradient $\nabla_{\pose(t)}\log p_{t}(\pose(t) | \cond )$ that intuitively guides the current pose to be more feasible to the task condition. To improve the sample quality, we simulate the reverse-time SDE via a Predictor-Corrector (PC) sampler~\cite{song2020score}\footref{footnote1}. 



\subsection{Masked Condition for Versatility}
\label{subsec:mask_condition}
To handle different applications and enhance the versatility, we design a hierarchical masking strategy to randomly mask the task condition $\cond$ while training $\score(\pose, t | \cond)$.
Concretely, we design a 3-level mask hierarchy to deal with human-pose-related tasks: $ M = M_{human} \odot M_{part} \odot M_{joint} $. $M_{human}$ indicates whether a condition $\cond$ is fully masked out with probability $p_h$ (result in $\O$ for unconditional pose prior $\data(\pose|\O)$). $M_{part}$ indicates whether each human body part in $\cond$ should be masked out with probability $p_{p}$. It facilitates recovery from occluded human body parts in the pose completion task. Practically, we think of humans as consisting of 5 body parts: 2 legs, 2 arms, torso. $M_{joint}$ indicates if we should randomly mask each human joint independently with probability $p_{j}$. This facilitates recovery from occluded human body joints in the pose completion task. We augment our model $\score(\pose, t | \cond)$ with $\cond = M \odot \cond$ during training.




\section{Experiments}
\label{sec:experiments}
In this section, we first provide summaries of the datasets and evaluation metrics we use and then elaborate on the implementation details. 
Then, we demonstrate the effectiveness and generalizability of GFPose under different problem settings, including pose estimation, pose denoising, pose completion, and pose generation.
Moreover, we ablate design factors of GFPose in the 3D human pose estimation task for in-depth analysis.

\subsection{Datasets and Evaluation Metrics}
\label{subsec:dataset_metric}
\paragraph{Human3.6M (H3.6M)~\cite{h36m_pami}} is a large-scale dataset for 3D human pose estimation, which consists of 3.6 million poses and corresponding images featuring 11 actors performing 15 daily activities from 4 camera views. Following the standard protocols, the models are trained on subjects $1$, $5$, $6$, $7$, $8$ and tested on subjects $9$, $11$. We evaluate the performance with the Mean Per Joint Position Error (MPJPE) measure following two protocols. \textit{Protocol \#1} computes the MPJPE between the ground-truth (GT) and the estimated 3D joint coordinates after aligning their root (mid-hip) joints. \textit{Protocol \#2} computes MPJPE after applying a rigid alignment between GT and prediction. For multi-hypothesis estimation, we follow the previous works~\cite{jahangiri2017generating, Li_2019_CVPR, li2020weakly, wehrbein2021probabilistic} to compute the MPJPE between the ground truth and the best 3D hypothesis generated by our model, denoted as minMPJPE.

\vspace{-1.25em}
\paragraph{MPI-INF-3DHP (3DHP)~\cite{mono-3dhp2017}} features more complex cases including indoor scenes, green screen indoor scenes, and outdoor scenes. We directly apply our model, which is trained on the H3.6M dataset, to 3DHP without extra finetuning to evaluate its generalization capability following the convention. We report the Percentage of Correctly estimated Keypoints (PCK) with a threshold of $150$ mm.


\subsection{Implementation Details}
\label{subsec:impl_details}
We use Stacked Hourglass network (SH) \cite{newell2016stacked} as our 2D human pose estimator for any downstream tasks that require 2D pose conditions, \eg, 3D human pose estimation. SH is pretrained on the MPII dataset \cite{andriluka20142d} and finetuned on the H3.6M dataset \cite{h36m_pami}. 
We set the time range of subVP SDE~\cite{song2020score} to $t \in [0, 1.0]$. 
To better demonstrate the effectiveness of the proposed pipeline, we choose a vanilla fully connected network~\cite{martinez_2017_3dbaseline} as the backbone of our score network. While many recent works take well-designed GNNs~\cite{lcn-pami,zeng2021learning} and Transformers~\cite{li2022mhformer,zheng20213d} as their backbone to enhance performance, we show GFPose can exhibit competitive results with a simple backbone. 
We use 2 residual blocks and set the hidden dimension of our score network to 1024 as in \cite{martinez_2017_3dbaseline}. 
We adopt exponential moving average with a ratio of 0.9999 and a quick warm start to stabilize the training process as suggested by~\cite{song2019generative}. 
We do not use additional data augmentation techniques commonly used in previous works, \eg, additional inputs (2d detection confidence, heatmaps, ordinal labels) and augmentation (horizontal flipping). We train GFpose with a batch size of 1000, learning rate $2e-4$, and Adam optimizer\cite{kingma2014adam}. 
Note that we can train a unified model for all downstream applications with a mixed condition masking strategy (when training together with 3D pose conditions, we add zeros to the last dimension of the 2D poses for a unified condition representation) or train separate models for each task with independent conditions and masking strategies. 
In the main paper, we report the performance of independently trained models, which slightly improve over the unified model on each task. We use ``HPJ-xxx" to denote the masking strategy used during training. \Eg, HPJ-010 means only the part level mask is activated with probability $0.1$. We use ``T" to denote the probability $1.0$. Please refer to the Supplementary for detailed settings of each task and results of the unified model.


\subsection{Monocular 3D Pose Estimation (2D$\rightarrow$3D)}
\subsubsection{Multi-Hypothesis}

\begin{table*}[!t]
\vspace{-0.4cm}
\center
\small
\setlength{\tabcolsep}{2pt}
\resizebox{\textwidth}{!}{
\renewcommand{\arraystretch}{1.1}
\setlength{\tabcolsep}{0.08cm}
\begin{tabular}{l c c c c c c c c c c c c c c c c}
\toprule
Protocol \#1 & Dire. & Disc. & Eat & Greet & Phone & Photo & Pose & Purch. & Sit & SitD & Smoke & Wait & WalkD & Walk & WalkT & Avg \\
\midrule
Martinez \etal~\cite{martinez_2017_3dbaseline} $(S=1)$  & 51.8 & 56.2 & 58.1 & 59.0 & 69.5 & 78.4 & 55.2 & 58.1 & 74.0 & 94.6 & 62.3 & 59.1 & 65.1 & 49.5 & 52.4 & 62.9 \\
Li \etal~\cite{li2020weakly} $(S=10)$ & 62.0 & 69.7 & 64.3 &  73.6 &  75.1 &  84.8 &  68.7 &  75.0 & 81.2 &  104.3 & 70.2 &  72.0 & 75.0 & 67.0 & 69.0 &  73.9 \\
Li \etal~\cite{Li_2019_CVPR} $(S=5)$ & 43.8 & 48.6&  49.1&  49.8&  57.6&  61.5&  45.9&  48.3 & 62.0 & 73.4 & 54.8&  50.6&  56.0&  43.4&  45.5 & 52.7 \\
Oikarinen \etal~\cite{oikarinen2021graphmdn} $(S=200)$ & 40.0& 43.2 &41.0 &43.4 &50.0& 53.6& 40.1 &41.4& 52.6& 67.3& 48.1& 44.2 &44.9 &39.5& 40.2& 46.2 \\
Sharma \etal~\cite{Sharma_2019_ICCV} $(S=200)$ & 37.8 &43.2& 43.0 &44.3 &51.1 &57.0& 39.7& 43.0& 56.3& 64.0& 48.1 &45.4& 50.4 &37.9 &39.9 &46.8 \\
Wehrbein \etal~\cite{wehrbein2021probabilistic} $(S=200)$ & 38.5 &42.5& 39.9& 41.7& 46.5& 51.6 &39.9 &40.8& 49.5 &56.8 &45.3 &46.4 &46.8& 37.8& 40.4 &44.3 \\
\hline
Ours $(S=10)$ & 39.9 & 44.6 &  40.2 &  41.3 &  46.7 &  53.6 &  41.9 &  40.4 & 52.1 & 67.1 & 45.7&  42.9&  46.1 & 36.5 &  38.0 & 45.1 \\
Ours $(S=200)$ & \textbf{31.7} & \textbf{35.4} &  \textbf{31.7} &  \textbf{32.3} &  \textbf{36.4} & \textbf{42.4} &  \textbf{32.7} &  \textbf{31.5} & \textbf{41.2} & \textbf{52.7} & \textbf{36.5} &  \textbf{34.0} &  \textbf{36.2} & \textbf{29.5} &  \textbf{30.2} & \textbf{35.6} \\

\midrule
Protocol \#2 & Dire. & Disc. & Eat & Greet & Phone & Photo & Pose & Purch. & Sit & SitD & Smoke & Wait & WalkD & Walk & WalkT & Avg \\
\midrule
 Martinez \etal~\cite{martinez_2017_3dbaseline} $(S=1)$ & 39.5 & 43.2 & 46.4 & 47.0 & 51.0 & 56.0 & 41.4 & 40.6 & 56.5 & 69.4 & 49.2 & 45.0 & 49.5 & 38.0 & 43.1 & 47.7 \\
 Oikarinen \etal~\cite{oikarinen2021graphmdn} $(S=200)$ & 30.8 & 34.7 & 33.6&  34.2 & 39.6 & 42.2& 31.0&  31.9&  42.9&  53.5 & 38.1 & 34.1 & 38.0&  29.6 & 31.1 & 36.3 \\
 Sharma \etal~\cite{Sharma_2019_ICCV} $(S=200)$  & 30.6  & 34.6 &  35.7  & 36.4 &  41.2 &  43.6  & 31.8  & 31.5  & 46.2 &  49.7 &  39.7 &  35.8 &  39.6 &  29.7  & 32.8 &  37.3 \\
 Wehrbein \etal~\cite{wehrbein2021probabilistic} $(S=200)$ & 27.9  & \textbf{31.4}  &29.7  &30.2 & 34.9 & 37.1 & 27.3 & 28.2 & 39.0 & 46.1 & 34.2 & 32.3  &33.6 & 26.1 & 27.5 & 32.4 \\
 \hline
 Ours $(S=200)$ & \textbf{26.4} & 31.5 & \textbf{27.2} & \textbf{27.4} & \textbf{30.3} & \textbf{36.1} & \textbf{26.8} & \textbf{26.0} & \textbf{38.4} & \textbf{45.8} & \textbf{31.2} & \textbf{29.2} & \textbf{32.2} & \textbf{23.1} & \textbf{25.8} & \textbf{30.5} \\
 Ours $(GT, S=200)$ & \underline{14.5} & \underline{17.3} & \underline{13.9} & \underline{16.3} & \underline{16.9} & \underline{15.2} & \underline{19.1} & \underline{22.3} & \underline{16.5} & \underline{16.6} & \underline{16.8} & \underline{16.6} & \underline{18.8} & \underline{14.0} & \underline{14.6} & \underline{16.9} \\
\bottomrule
\end{tabular}}
\caption{Pose estimation results on the H3.6M dataset. We report the minMPJPE(mm) under Protocol\#1 (no rigid alignment) and Protocol\#2 (with rigid alignment). $S$ denotes the number of hypotheses. $GT$ indicates the condition is ground truth 2D poses.}
\label{table:h36m}
\end{table*}


\vspace{-0.5em}
\paragraph{Results on H3.6M}
Based on the conditional generative formulation, it is natural to use GFPose to generate multiple 3D poses conditioned on a 2D observation. Following previous works~\cite{Sharma_2019_ICCV, wehrbein2021probabilistic}, we produce $S$ 3D pose estimates for each detected 2D pose and report the minMPJPE between the GT and all estimates. As shown in Table \ref{table:h36m}, when 200 samples are drawn, our method outperforms the SOTAs~\cite{oikarinen2021graphmdn, Sharma_2019_ICCV, wehrbein2021probabilistic} by a large margin.
When only 10 samples are drawn, GFPose can already achieve comparable performance to the SOTA methods~\cite{wehrbein2021probabilistic} with 200 samples, indicating the high quality of the learned pose priors.

\vspace{-1.25em}
\paragraph{Results on 3DHP}
We evaluate GFPose on the 3DHP dataset to assess the cross-dataset generalization. Neither the 2D detector nor the generative model is finetuned on 3DHP. As shown in Table \ref{table:mpi} , our method achieves consistent performance across different scenarios and outperforms previous methods~\cite{wehrbein2021probabilistic,Li_2019_CVPR} even if ~\cite{Li_2019_CVPR} uses GT 2D joints. GFPose also surpasses~\cite{li2020weakly}, although it is specifically designed for transfer learning. 

\begin{table}
    \resizebox{0.46\textwidth}{!}{
		\renewcommand{\arraystretch}{1}
        \begin{tabular}{l c c c c}
            \toprule
            Method & GS & noGS & Outdoor & ALL PCK \\
            \midrule
            Li \etal~\cite{Li_2019_CVPR} & 70.1 & 68.2 & 66.6 & 67.9 \\
            Wehrbein \etal~\cite{wehrbein2021probabilistic} & 86.6 & 82.8 & 82.5 & 84.3 \\
            Li \etal~\cite{li2020weakly} & 86.9 & 86.6 & 79.3 & 85.0 \\
            Ours & \textbf{88.4} & \textbf{87.1} & \textbf{84.3} & \textbf{86.9} \\
            \bottomrule
        \end{tabular}
    }
    \vspace{-0.2cm}
    \caption{Pose estimation results on the 3DHP dataset. 200 samples are drawn. ``GS" represents the ``Green Screen" scene. Our method outperforms~\cite{Li_2019_CVPR, wehrbein2021probabilistic, li2020weakly}, although~\cite{li2020weakly} is specifically designed for domain transfer.}
    \label{table:mpi}
    \vspace{-0.5em}
\end{table}


\subsubsection{Single-Hypothesis}
We further evaluate GFPose under a single-hypothesis setting, \ie, only 1 hypothesis is drawn. Table \ref{table:cmp_deterministic} reports the MPJPE(mm) of current probabilistic (one-to-many mapping) and deterministic (one-to-one mapping) methods. Our method outperforms the SOTA probabilistic approaches by a large margin under the single hypothesis setting. This shows that GFPose can better estimate the likelihood . 
Moreover, we find GFPose can also achieve comparable results to the SOTA deterministic methods~\cite{zeng2020srnet}, even with a plain fully-connected network. In contrast, SOTA deterministic methods~\cite{ci2019optimizing, zeng2020srnet} use specifically designed architectures, \eg, Graph Networks~\cite{ci2019optimizing, zeng2020srnet} or a stronger 2D pose estimator (CPN~\cite{chen2018cascaded})~\cite{pavllo20193d, zeng2020srnet} to boost performance. This in turn shows the great potential of GFPose. We believe the improvement of GFPose comes from the new probabilistic generative formulation which mitigates the risk of over-fitting to the mean pose as well as the powerful gradient field representation.

To further verify this, we use the backbone of GFPose and train it in a deterministic manner. We test it on different 2D pose distributions and compare it side-by-side with our GFPose. MPJPE measurements are reported in Table \ref{table:cmp_sup}. We can find that GFPose (P) consistently outperforms its deterministic counterpart (D) on both same- and cross-distribution tests, demonstrating a higher accuracy and robustness. Although it is hard to control all the confounding factors (\eg, finding the best hyperparameter sets for each setting), arguably, learning score is a better alternative to learning deterministic mapping for 3D pose estimation.




\begin{table}
    \centering
    \small
		\renewcommand{\arraystretch}{1}
        \begin{tabular}{l l c}
            \toprule
            Type & Method & MPJPE(mm) \\
            \midrule
            \multirow{5}{*}{Probabilistic} & Li \etal~\cite{li2020weakly} & 80.9 \\
            & Li \etal~\cite{Li_2019_CVPR} & 62.9 \\
            & Wehrbein \etal~\cite{wehrbein2021probabilistic} & 61.8 \\
            & Oikarinen \etal~\cite{oikarinen2021graphmdn} & 59.2 \\
            & Ours & \textbf{51.0} \\
            \midrule
            \multirow{4}{*}{Deterministic} & Martinez \etal~\cite{martinez_2017_3dbaseline} & 62.9 \\
            & Ci \etal~\cite{ci2019optimizing} & 52.7 \\
            & Pavllo \etal~\cite{pavllo20193d} & 51.8 \\
            & Zeng \etal~\cite{zeng2020srnet} & \textbf{49.9} \\
            \bottomrule
        \end{tabular}
    \caption{Single-hypothesis results on the H3.6M dataset. We report MPJPE(mm) under Protocol \#1. The upper body of the table lists the SOTA probabilistic methods. The lower body of the table lists the SOTA deterministic methods. We demonstrate that the probabilistic method can also achieve competitive results under the single-hypothesis setting for the first time.}
    \label{table:cmp_deterministic}
\end{table}

\begin{table}
    \centering
    \small
    \resizebox{0.45\textwidth}{!}{
		\renewcommand{\arraystretch}{1}
        \begin{tabular}{l c c c c}
            \toprule
            Train/Test & GT & DT & GT+$\mathcal{N}$(0, 25) & DT+$\mathcal{N}$(0, 25) \\
            \midrule
            P (GT) & \textbf{35.6} & \textbf{55.7} & \textbf{72.2}  & \textbf{81.2}  \\
            D (GT) & 41.9 & 61.6  & 77.9 & 89.2 \\
            \midrule
            P (DT) & \textbf{38.9} & \textbf{51.0} & \textbf{61.1} & \textbf{69.0} \\
            D (DT) & 46.9 & 57.0 & 64.4 & 72.1 \\
            \bottomrule
        \end{tabular}
    }
    \caption{Side-by-side comparison between the probabilistically trained score model (P) and the deterministically trained counterpart (D). Models are trained on GT 2D poses (GT) or Stack Hourglass detected 2D poses (DT) and test with different 2D pose distributions. $\mathcal{N}$ indicates Gaussian noise. Only one sample is drawn from P. We report MPJPE(mm) under Protocol\#1 on H3.6M.}
    \label{table:cmp_sup}
\end{table}

\subsection{Pose Completion (Incomplete 2D/3D $\rightarrow$ 3D)}
2D pose estimation algorithms and MoCap systems often suffer from occlusions, which result in incomplete 2D pose detections or 3D captured data. As a general pose prior model, GFPose can also help to recover an intact 3D human pose from incomplete 2D/3D observations. Due to the condition masking strategy, fine-grained completion can be done at either the joint level or body part level.

\begin{table}
    \centering
    \small
		\setlength{\tabcolsep}{0.4cm}
        \begin{tabular}{l c c}
            \toprule
            Occ. Body Parts & Ours &  Li \etal~\cite{Li_2019_CVPR} \\
            \midrule
            1 Joint & \textbf{37.8} & 58.8 \\
            2 Joints & \textbf{39.6} & 64.6 \\
            \midrule
            2 Legs & 53.5 & -\\
            2 Arms & 60.0 & -\\
            Left Leg + Left Arm & 54.6 & - \\
            Right Leg + Right Arm  & 53.1 & - \\
            \bottomrule
        \end{tabular}
    \vspace{-0.4em}
    \caption{Recover 3D pose from partial 2D observation. We train two separate models with masking strategy HPJ-001 and HPJ-020 for random missing joints and body parts, respectively. We report minMPJPE(mm) with 200 samples under Protocol \#1. Our HPJ-001 model significantly outperforms~\cite{Li_2019_CVPR} even though they train 2 models to deal with different numbers of missing joints while we only use one model to handle varying numbers of missing joints.}
    \label{table:complete_2d}
\end{table}

\begin{table}
    \centering
    \small
 		\setlength{\tabcolsep}{0.7cm}
        \begin{tabular}{l c c}
            \toprule
            Occ. Body Parts & $S=1$ & $S=200$ \\
            \midrule
            Right Leg & \textbf{13.0} & \textbf{5.2} \\
            Left Leg & 14.3 & 5.8\\
            Left Arm & 25.5 & 9.4 \\
            Right Arm & 22.4 & 8.9 \\
            \bottomrule
        \end{tabular}
    \vspace{-0.4em}
    \caption{Recover 3D pose from partial 3D observation. We report minMPJPE(mm) under Protocol \#1. $S$ denotes the number of samples.}
    \label{table:complete_3d}
    \vspace{-1.0em}
\end{table}

\paragraph{Recover 3D pose from partial 2D observation}
We first evaluate GFPose given incomplete 2D pose estimates. This is a very common scenario in 2D human pose estimation. Body parts are often out of the camera view and body joints are often occluded by objects in the scene.
We train two separate models conditioning on 2D incomplete poses with masking strategy HPJ-001 and HPJ-020 to recover from random missing joints and body parts respectively. We draw 200 samples and report minMPJPE in Table \ref{table:complete_2d}. Our method outperforms~\cite{Li_2019_CVPR} by a large margin even though they train 2 models to deal with different numbers of missing joints. Our method also shows a smaller performance drop when the number of missing joints increases (1.8mm vs. 5.8mm), which validates the robustness of GFPose. In addition, we provide more numerical results on recovering from occluded body parts. This indicates more severe occlusion (6 joints are occluded) where less contextual information can be explored to infer joint locations. GFPose still shows compelling results. In most cases, it outperforms ~\cite{Li_2019_CVPR} although our method recovers from 6 occluded joints while ~\cite{Li_2019_CVPR} recovers from only 1 occluded joint. This indeed demonstrates that GFPose learns a strong pose prior.

\begin{figure}[!t]
    \centering
    \includegraphics[width=0.99\linewidth]{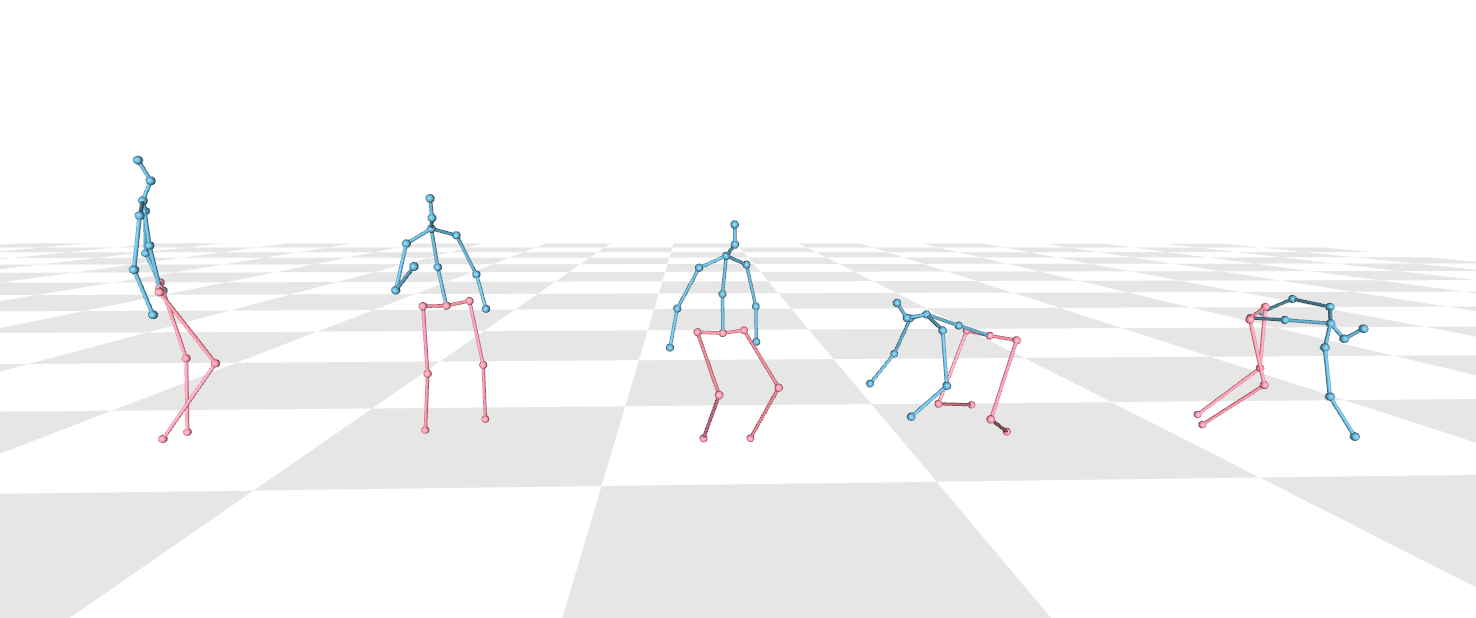}
    \vspace{-0.6em}
    \caption{The \textcolor[RGB]{255,143,163}{lower body} completed by GFPose given only the \textcolor[RGB]{97,165,194}{upper body}. One sample is drawn for each pose.}
    \vspace{-1.5em}
    \label{fig:viz_complete}
\end{figure}

\paragraph{Recover 3D pose from partial 3D observation}
Fitting to partial 3D observations has many potential downstream applications, \eg, completing the missing legs for characters in VR applications (metaverse). We show that GFPose can be directly used to recover missing 3D body parts given partial 3D observations.
In this task, we train GFPose with 3D poses as conditions. We adopt the masking strategy HPJ-020.
Table \ref{table:complete_3d} shows the minMPJPE. We can find that GFPose does quite well in completing partial 3D poses. When we sample 200 candidates, minMPJPE reaches a fairly low level. 
If we take a further look, we can see that legs are more difficult to recover than arms. We believe it is caused by the greater freedom of arms. Compared to legs, the plausible solution space of arms is not well constrained given the positions of the rest body. We show some qualitative results of completing lower body given upper body in Figure \ref{fig:viz_complete}.

\vspace{-0.1em}
\subsection{Denoise MoCap Data (Noisy 3D $\rightarrow$ Clean 3D)}
\vspace{-0.2em}
3D human poses captured from vision-based algorithms or wearable devices often suffer from different types of noises, such as jitters or drifts. Due to the denoising nature of score-based models, it is straightforward to use GFPose to denoise MoCap Data. Following previous works~\cite{tiwari2022pose}, we evaluate the denoising effect of GFPose on a noisy MoCap dataset. Concretely, we add uniform noise and Gaussian noise with different intensities to the test set of H3.6M, and evaluate GFPose on this ``noisy H3.6M" dataset. 
We train GFPose with a masking strategy HPJ-T00 (always activate human level mask with a probability 1.0). At test time, we condition GFPose on $\O$ and replace the initial state $\pose(T)$ with the noisy pose. Generally, we choose a smaller start time for smaller noise.
Table \ref{table:mocap_denoise} reports the MPJPE (mm) under Protocol \#2 before and after denoising. We find that GFPose can effectively handle different types and intensities of noise. For Gaussian noise with a small variance of 25, GFPose can improve the pose quality by about 24\% in terms of MPJPE. For Gaussian noise with a large variance of 400, GFPose can improve the pose quality by about 49\%. We also observe consistent improvement in uniform noise. In addition, we report the denoising results on a clean dataset, as shown in the first row of Table \ref{table:mocap_denoise}. A good denoiser should keep minimum adjustment on the clean data. We find our GFPose introduces moderate extra noise on the clean data.

Most previous works~\cite{tiwari2022pose, rempe2021humor, pavlakos2019expressive, davydov2022adversarial} learn the pose priors in the SMPL~\cite{loper2015smpl} parameter space. While our method learns pose priors on the joint locations. It is hard to directly compare the denoising effect between our method and previous works. Here, we list the denoising performance of a most recent work Pose-NDF\cite{tiwari2022pose} as a reference. According to~\cite{tiwari2022pose}, the average intensity of introduced noise is 93.0mm, and the per-vertex error after denoising is 79.6mm. Note that they also leverage additional temporal information to enforce smoothness.

\begin{table}
    \resizebox{0.46\textwidth}{!}{
        \setlength{\tabcolsep}{0.4cm}
        \begin{tabular}{l l c}
            \toprule
            Input Data & MPJPE (before/ after) & Start $T$ \\
            \midrule
            GT  & \textbf{0} / 14.7 & 0.05 \\  
            \midrule
            GT + $\mathcal{N}(0, 25)$ & 33.1 / \textbf{25.0} & 0.05 \\
            GT + $\mathcal{N}(0, 100)$ & 65.5 / \textbf{42.8} & 0.05 \\
            GT + $\mathcal{N}(0, 400)$ & 126.0 / \textbf{64.6} & 0.1 \\
            GT + $\mathcal{U}(25)$ & 49.1 / \textbf{32.8} & 0.05 \\
            GT + $\mathcal{U}(50)$ & 96.2 / \textbf{50.9} & 0.1 \\
            GT + $\mathcal{U}(100)$ & 178.2 / \textbf{89.4} & 0.1 \\
            \bottomrule
        \end{tabular}
    }
    \vspace{-0.2cm}
    \caption{Denoising results on H3.6M dataset. We report MPJPE (mm) under Protocol \#2. $\mathcal{N}$ and $\mathcal{U}$ denote Gaussian and uniform noise respectively. $T$ denotes the start time of RSDE.}
    \label{table:mocap_denoise}
    \vspace{-0.5cm}
\end{table}

\subsection{Pose Generation (Noise $\rightarrow$ 3D)}
Getting annotated 3D pose is expensive. We show that GFPose can also be used as a pose generator to produce diverse and realistic 3D poses. 
In this task, we train GFPose with a maksing strategy HPJ-T00. At test time, we condition GFPose on $\O$ to generate random poses from $\data(\pose)$.
To assess the diversity and realism of the generated poses, we train a deterministic pose estimator on the generated data and evaluate it on the test set of H3.6M. Note that the pose estimator gets satisfactory performance only when the generated data are both diverse and realistic. It is in fact a strict metric. We sample 177,200 poses from GFPose to compose the synthetic dataset. While the original H3.6M training set consists of 1,559,752 poses. 
Table \ref{table:pose_generation} shows that the pose estimator can achieve moderate performance when trained only with the synthetic data. This demonstrates GFPose can draw diverse and realistic samples. 

In addition, we find that the generated data can also serve as an augmentation to the existing H3.6M training set and further benefit the training of a deterministic pose estimator.  We experiment with different scales of real data and synthetic data to simulate two typical scenarios where MoCap data are scarce or abundant. 
When the size of Mocap data is small (2,438 samples), increasing the size of synthetic data will continuously improve performance.
When the size of Mocap data is large (1,559,752 samples), complementing it with a small proportion of synthetic data can still boost the performance. This further demonstrates the value of GFPose as a pose generator.

\begin{table}[t]
    \centering
    \small
        \begin{tabular}{l l c}
            \toprule
            Mocap Data (H3.6M) & Synthetic Data (GFPose) & MPJPE  \\
            \midrule
            2,438 & 0 & 66.1  \\
            2,438 & 1,219 & 62.0 \\
            2,438 & 2,438 & 60.0\\
            2,438 & 177,200 & \textbf{58.1} \\
            \midrule
            1,559,752 & 0 & 54.3\\
            1,559,752 & 177,200 & \textbf{53.4}  \\
            \midrule
            0 & 177,200 & \textbf{58.1}  \\
            \bottomrule
        \end{tabular}
     \vspace{-0.5em}
    \caption{Augment training data with sampled poses from GFPose. We train separate deterministic pose estimators with different combinations of real / synthetic data. The number of 1,559,752 is the size of the H3.6M training set. The number of 2,438 indicates sampling the original H3.6M training set every 640 frames. MPJPE(mm) under Protocol \#1 is reported.}
    \label{table:pose_generation}
    \vspace{-0.5em}
\end{table}

\subsection{Ablation Study}

We ablate different factors that would affect the performance of our model on the 3D pose estimation task, including the number of sampling steps, and the depth and width of the network. As shown in Table \ref{table:ablation}, more sampling steps or deeper or wider networks do not improve the reconstruction accuracy, but at the expense of greatly reducing the inference speed. We leave the ablation of different condition masking strategies to the Supplementary. 

\begin{table}
    \centering
    \small
 		\setlength{\tabcolsep}{0.35cm}
        \begin{tabular}{c c c c c}
            \toprule
            Steps & Blocks & Hidden Dim & MPJPE & FPS  \\
            \midrule
            1k & 2 & 1024 & \textbf{35.6} & 240 \\
            \hline
            0.5k & 2 & 1024 & 36.1 & \textbf{594} \\
            2k & 2 & 1024 & 35.7 & 107\\
            1k & 3 & 1024 & 37.2 & 154\\
            1k & 4 & 1024 & 37.1 & 123 \\
            1k & 2& 512 & 44.3 & 461 \\
            1k & 2& 2048  & 37.6 & 79 \\
            1k & 2& 4096  & 37.5 & 35 \\
            \bottomrule
        \end{tabular}
   \vspace{-0.5em}
    \caption{Ablation Study. We draw 200 samples for each model and report minMPJPE(mm) under Protocol \#1. We also compare the inference speed to draw one sample on a NVIDIA 2080Ti GPU.}
    \label{table:ablation}
    \vspace{-1.0em}
\end{table}



\section{Conclusion and Discussion}
\label{sec:conclusion}

We introduce GFPose, a versatile framework to model 3D human pose prior via denoising score matching. GFPose incorporates pose prior and task-specific conditions into gradient fields for various applications. We further propose a condition masking strategy to enhance the versatility. We validate the effectiveness of GFPose on various downstream tasks and demonstrate compelling results.

\noindent\textbf{Limitation} Although GFPose has shown great potential in many applications, the reverse sampling process requires repeated model inferences. The overall inference time is proportional to the number of sampling steps,
which may limit the use of very large deep networks in real-time scenarios.


{\small
\bibliographystyle{ieee_fullname}
\bibliography{egbib}
}

\clearpage

\appendix





\section{Details of subVP SDE}
In this work, we use the subVP SDE proposed in ~\cite{song2020score} to perturb the 3D pose data. Formally, 

\begin{equation}
\begin{aligned}
    d\pose = - \frac{1}{2} \beta(t) \pose dt + \sqrt{ \beta (t) ( 1 -  e^{-2  \int_{0}^{t} \beta (s) \, ds } ) }   d\mathbf{w},
\end{aligned}
\end{equation}
where $\pose \in \R^{J \times 3}$ denotes the 3D human pose, $\beta (t)$ denotes the noise scale at timestep $t$.
The drift coefficient of $\pose (t)$ is $\mathbf{f}(\pose, t) = - \frac{1}{2} \beta(t) \pose $. The diffusion coefficient is $g(t) = \sqrt{ \beta (t) ( 1 -  e^{-2  \int_{0}^{t} \beta (s) \, ds } ) }$.
$t \in [0, 1]$ is a continuous variable. We adopt the linear scheduled noise scales:

\begin{equation}
\begin{aligned}
    \beta (t) = \beta(0) + t \left( \beta(1) - \beta(0) \right).
\end{aligned}
\end{equation}
We empirically set the minimum and maximum noise scale $\beta(0)$ and $\beta(1)$ to $0.1$ and $20.0$, respectively.

\section{Closed Form of Loss Function}
Because the drift coefficient $\mathbf{f}(\pose, t)$ is affine, according to \cite{song2020score}, the transition kernel $p_{0t}(\pose(t) \mid \pose(0), \cond)$ in the loss function (Eq.~\ref{eq: SDE score matching} in the main text) is always a Gaussian distribution, where the mean and variance can be obtained in the closed form:

\begin{equation}
\begin{aligned}
    \mathcal{N} \left( \pose(t);\pose(0) e^{- \frac{1}{2}  \int_{0}^{t} \beta (s) \, ds }, \left[ 1 -  e^{-\int_{0}^{t} \beta (s) \, ds }\right]^2 \mathbf{I} \right).
\end{aligned}
\label{eq:trans_kernal}
\end{equation}

Following~\cite{song2019generative}, we choose the weighting function $\lambda(t) = \sigma(t)^2$. Thus, the loss function can be written as:
\begin{equation}
\begin{aligned}
    \loss & = {} \E_{\mathcal U(t;0, 1)} \left[ \lambda(t) \norm{\score \left( \pose(t), t, \cond \right) + \frac{\pose(t) - \mu}{\sigma^2} }_2^2 \right] \\
      & = {} \E_{\mathcal U(t;0, 1)} \left[\norm{\sigma(t) \score\left(\pose(t), t, \cond\right) + \z }_2^2 \right],
\end{aligned}
\label{eq:closed_loss}
\end{equation}
where $\z \sim \mathcal{N}(0, 1)$ is random noise. $\score \left( \pose(t), t, \cond \right)$ is the score network we would like to learn. According to Eq. \ref{eq:trans_kernal}, we can get $\sigma(t) = 1 -  e^{-\int_{0}^{t} \beta (s) \, ds}$.

During training, we uniformly sample the time variable $t$ from $[0, 1]$ and sample a noise vector $\z \in \R^{J \times 3}$ from a standard normal distribution. Then we perturb the ground-truth 3D human pose $\pose(0)$ according to Eq.~\ref{eq:trans_kernal} to get the noisy 3D pose $\pose(t)$:

\begin{equation}
    \pose(t) = \pose(0) e^{- \frac{1}{2}  \int_{0}^{t} \beta (s) \, ds } + \z \cdot \left( 1 -  e^{-\int_{0}^{t} \beta (s) \, ds } \right). 
\end{equation}
Then we can compute the loss according to Eq.~\ref{eq:closed_loss} to train the score network.

\section{Network Architecture}
\begin{figure*}[!t]
    \centering
    \includegraphics[width=0.97\linewidth]{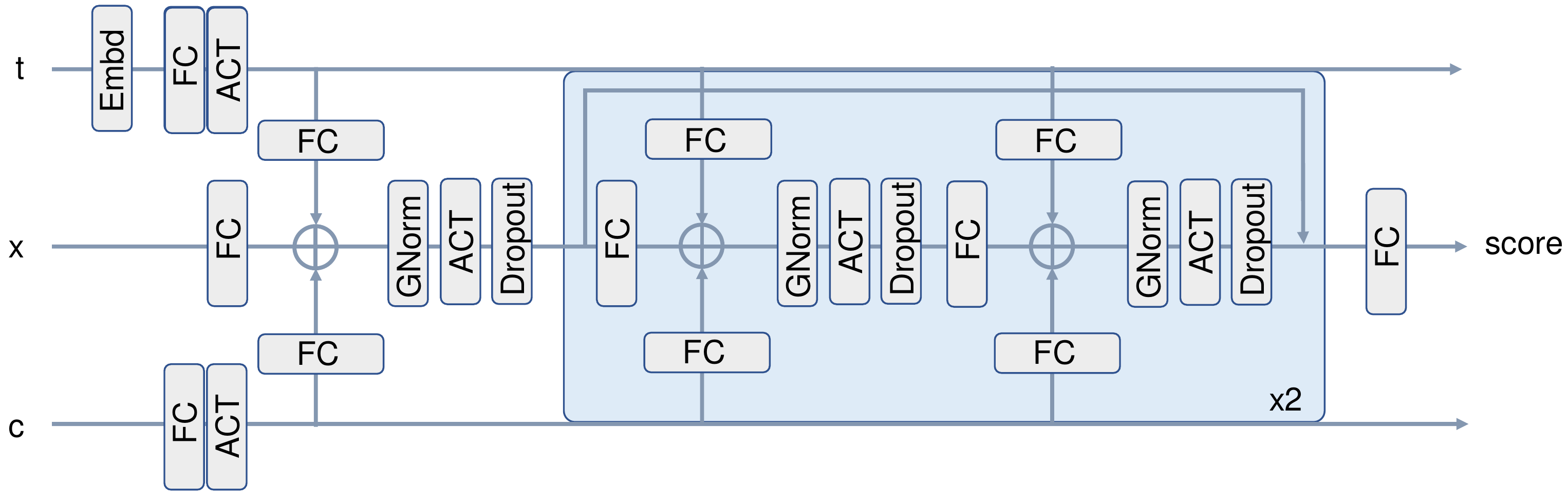}
    \caption{Architecture of the score network $\score$. It is a plain fully connected network consisting of 2 residual blocks.  $\pose \in \R^{J \times 3}$ denotes noisy 3D poses. $\cond$ denotes different task conditions, \eg detected 2D poses. $t$ denotes the timestep. $\oplus$ denotes the sum operator.  }
    \label{fig:supp_network}
\end{figure*}

Fig.~\ref{fig:supp_network} shows the architecture of our score network $\score$. It is a simple fully-connected network with a structure similar to \cite{martinez_2017_3dbaseline}. Following \cite{martinez_2017_3dbaseline}, we set the hidden dimension of all FC layers to $1024$. We adopt group normalization \cite{wu2018group} with the number of groups set to $32$. We use SiLU \cite{hendrycks2016gaussian} as nonlinear activation and set the dropout rate to $0.25$.

\section{Pose Sampling}
\label{sec:supp_sampling}
To sample human poses from the learned pose prior $\data(\pose|\cond)$, we need to solve the reverse-time SDE (Eq.~\ref{eq:reverse_sde} in the main text). Following~\cite{song2020score}, we simulate the RSDE via the Predictor-Corrector (PC) sampler. We use the Euler-Maruyama solver as the predictor and Identity as the corrector. We set the number of sampling steps $N=1000$, start time $T=1.0$ and the end time $eps=1e-3$. Please refer to Alg.~\ref{alg:sampling} for the detailed sampling process.

\section{Detailed Settings for Different Tasks}
\label{sec:supp_task}
In this section, we detail how we train and use our model for each task described in the main text. The difference lies in the types of conditions, masking strategy and sampling process. 


\paragraph{Monocular 3D Human Pose Estimation}
During training, GFPose conditions on the detected 2D pose without any condition masking strategy, \ie, HPJ-000. At inference, we sample a noise vector $\pose(T) \in \R^{J \times 3}$ from a standard normal distribution $\mathcal{N} (0, 1)$, and gradually adjust it according to the standard sampling process described in Alg.~\ref{alg:sampling}.

\begin{algorithm}[t]
\caption{Sampling from $\data(\pose|\cond)$}
\label{alg:sampling}
\begin{algorithmic}[1]
\Require learned $\score$; sampling step $N$; condition $c$; start time $T$; end time $eps$
\State $\mathbf{f}(\pose, t) \gets - \frac{1}{2} \beta(t) \pose$
\State $g(t) \gets \sqrt{ \beta (t) ( 1 -  e^{-2  \int_{0}^{t} \beta (s) \, ds } ) }$
\State $dt \gets \frac{1}{N}$
\State $\pose_N \sim \mathcal{N}(0, 1)$
\For{$i \gets N-1$ to $0$}
    \State $t \gets eps + (T - eps) \cdot \frac{i+1}{N} $
    \State $\pose_{i}' \gets \pose_{i+1} - \left[ \mathbf{f}(\pose_{i+1}, t) - g(t)^2 \score \left( \pose_{i+1}, t, \cond \right) \right] dt$
    \State $\z \sim \mathcal{N}(0, 1)$
    \State $\pose_{i} \gets \pose_{i}' + g(t) \sqrt{dt} \z$
\EndFor
\State \Return $\pose_{0}$
\end{algorithmic}
\end{algorithm}

\paragraph{Pose Completion (2D $\rightarrow$ 3D)}
During training, GFPose conditions on the detected 2D pose with masking strategy HPJ-001 and HPJ-020 for missing joint and part completion, respectively. 
Here, we explore three sampling approaches. The first is the standard conditional inference process as described in Alg.~\ref{alg:sampling}. The second is the imputation approach proposed in \cite{song2020score}. We also try combining these two inference techniques as the third approach.  In this task, we adopt the combination approach as it performs best. In Sec.~\ref{sec:supp_imputation}, we will further compare different inference approaches.

\paragraph{Pose Completion (3D $\rightarrow$ 3D)}
During training, GFPose conditions on the gt 3D pose with a masking strategy HPJ-020 for missing body part completion. We also adopt the combination approach during sampling.


\paragraph{Denoising Mocap Data}
In this task, GFPose is trained with a masking strategy HPJ-T00, \ie, human level mask is always activated and the condition is always zero. To use this model for denoising, we condition the model on $\O$ and start the sampling process from a noisy 3D pose $\pose(T) \in \R^{J \times 3}$ instead of a noise vector. We set the start time $T$ of reverse-time SDE to a small value of $0.05$ or $0.1$ instead of the default value $1.0$, as shown in Table~\ref{table:mocap_denoise} in the main text. Intuitively, a smaller start time notifies the score network $\score$ that we are not starting from pure noise, so a small adjustment is sufficient.

\paragraph{Pose Generation}
For generation, GFPose is trained with a masking strategy HPJ-T00. During inference, GFPose conditions on $\O$ and gradually adjusts the noise vector $\pose(T) \in \R^{J \times 3}$ sampled from a standard normal distribution $\mathcal{N} (0, 1)$ to generate realistic and diverse 3D poses. 

\section{Train A Unified Model for Various Tasks}
Instead of training separate models for different tasks, we can also train one unified model for all tasks. Concretely, we active all three levels of masks during training. There are two different choices to train such a unified model. 1) We only condition on the detected 2D pose, \ie \ $\cond \subset \{ \pose_{2d} \} $. We call this model ``U2D". 2) We condition on the detected 2D pose $0.9$ and the 3D pose with a probability of $0.1$, \ie \ $\cond \subset \{ \pose_{2d}, \pose_{3d} \} $. We call this model ``U3D". Note that the 3D condition shares the same masking strategy as the 2D condition. This helps the model to explicitly establish the relationship between different 3D body parts. We quantitatively study the unified model in the following.

\begin{table}[t]
    \centering
    \small
        \begin{tabular}{c c c c c c}
            \toprule
            $p_h$ & $p_p$ & $p_j$ & 2D & 3D & minMPJPE ($S=1 / 200$) \\
            \midrule
            0.0 & 0.0 & 0.0 & \cmark & \xmark & 51.0 / 35.6  \\
            \hdashline
            0.1 & 0.0 & 0.0 & \cmark & \xmark & 50.8 /35.8   \\
            0.0 & 0.1 & 0.0 & \cmark & \xmark & \textbf{50.5} / \textbf{35.1} \\
            0.0 & 0.0 & 0.1 & \cmark & \xmark & 51.7 / 36.4 \\
            0.1 & 0.1 & 0.0 & \cmark & \xmark & 51.0 / 36.0 \\
            0.0 & 0.1 & 0.1 & \cmark & \xmark & 51.1 / 35.5 \\
            \hdashline
            0.1 & 0.1 & 0.1 & \cmark & \xmark & 52.3 / 36.7 \\
            0.1 & 0.2 & 0.1 & \cmark & \xmark & 52.4 / 36.5 \\
            0.1 & 0.1 & 0.1 & \cmark & \cmark & \textbf{51.3} / \textbf{35.9}  \\
            0.1 & 0.2 & 0.1 & \cmark & \cmark & 51.6 / 36.4  \\
            \bottomrule
        \end{tabular}
    \caption{Effects of condition masking strategies on monocular 3D human pose estimation. We report minMPJPE(mm) on H3.6M dataset under protocol \#1. S denotes the number of samples. $p_h$, $p_p$ and $p_j$ respectively denote the probability of activating human-, part- and joint- level masks.}
    \label{table:supp_cond_masking}
\end{table}

\paragraph{Monocular 3D Pose Estimation} We first ablate condition masking strategies on monocular 3D human pose estimation. Table \ref{table:supp_cond_masking} reports the minMPJPE(mm) for different models on H3.6M dataset.  We can find that training with human- or part- level masks slightly boosts the performance of pose estimation ($\sim$0.5mm). Training with the mixed masking strategy, \ie \ the unified model ``U2D" (3rd and 4th row from last) causes a slight drop ($\sim$1mm) in reconstruction accuracy. Further incorporating 3D pose  (1st and 2nd row from last, ``U3D") can benefit the pose estimation task. 

\paragraph{Pose Completion}
We compare the two unified models ``U2D" and ``U3D" on the pose completion task. From Table \ref{table:supp_complete_2d}, we can observe that the unified models perform slightly worse than the specifically trained model when recovering occluded body parts. However, they achieve competitive results as the specifically trained model when recovering occluded random joints and significantly outperforms the SOTA~\cite{Li_2019_CVPR}.
From Table \ref{table:supp_complete_3d}, we can find that the unified model ``U3D" performs quite well when recovering from partial 3D observations. Note that ``U2D" cannot directly apply to this task unless through the imputation technique introduced in the next section.

\paragraph{Pose Denoising}
Table~\ref{table:supp_denoising} shows that the unified model performs slightly better on small noise intensities while the specifically trained model (HPJ-T00) performs slightly better on large noise intensities. 

\paragraph{Pose Generation}
We train 3 deterministic pose estimators on the synthetic poses generated by HPJ-T00, ``U2D" and ``U3D", respectively. Table~\ref{table:supp_generation} shows that the specifically trained model HPJ-T00 can generate the best quality poses. Unified models also show competitive results.

\paragraph{Summary}
We can trade very little performance loss for a versatile unified model ``U3D".

\begin{table}
    
    \centering
    \small
        \begin{tabular}{l c c c c}
            \toprule
            Occ. Parts & Sep. & U2D & U3D &  Li \etal~\cite{Li_2019_CVPR} \\
            \midrule
            1 Joint & 37.8 & \textbf{37.5} & 37.6 & 58.8 \\
            2 Joints & 39.6 & 39.8 & \textbf{39.5} & 64.6 \\
            \midrule
            2 Legs & \textbf{53.5} & 54.9 & 53.8 & -\\
            2 Arms & \textbf{60.0} & 62.1 & 60.4 & -\\
            Left Leg + Left arm & \textbf{54.6} & 56.4 & 55.2 & - \\
            Right leg + Right arm  & \textbf{53.1} & 54.3 & 53.5 & - \\
            \bottomrule
        \end{tabular}
    \caption{Recover 3D pose from partial 2D observation. We compare the unified models (``U2D" and ``U3D") and the model specifically trained for this task (Sep. here means HPJ-001 and HPJ-020, reported in the main text. Please refer to Section \ref{sec:supp_task}). We report minMPJPE(mm) on H3.6M dataset under protocol \#1. 200 samples are drawn.}
    \label{table:supp_complete_2d}
\end{table}

\begin{table}
    \centering
    \small
 	\setlength{\tabcolsep}{0.4cm}
        \begin{tabular}{l c c}
            \toprule
            Occ. Parts & Sep. & U3D \\
            \midrule
            Right Leg &  \textbf{5.2} & 5.6 \\
            Left Leg &  \textbf{5.8} & 5.9 \\
            Left Arm &  \textbf{9.4} & 9.7 \\
            Right Arm &  \textbf{8.9} & 9.0\\
            \bottomrule
        \end{tabular}
    \caption{Recover 3D pose from partial 3D observation. We compare the unified models (``U2D" and ``U3D") and the model specifically trained for this task (Sep., reported in the main text. Please refer to Section \ref{sec:supp_task} for details) on H3.6M dataset and report minMPJPE(mm) under protocol \#1. 200 samples are drawn.}
    \label{table:supp_complete_3d}
\end{table}

\begin{table}
    \resizebox{0.46\textwidth}{!}{
        \begin{tabular}{l c c c c c}
            \toprule
            Noisy Data & Base & Sep. & U2D & U3D & Start $T$ \\
            \midrule
            GT  & 0 & 14.7 & \textbf{13.8} & 14.0 & 0.05  \\  
            \midrule
            GT + $\mathcal{N}(0, 25)$ & 33.1 & 25.0 & 25.1 & \textbf{24.8} & 0.05 \\
            GT + $\mathcal{N}(0, 100)$ & 65.5 & \textbf{42.8} & 44.0 & 43.5 & 0.05 \\
            GT + $\mathcal{N}(0, 400)$ & 126.0 & \textbf{64.6} & 65.5 & 65.2 & 0.1 \\
            GT + $\mathcal{U}(25)$ & 49.1 & \textbf{32.8} & 33.5 & 33.0 & 0.05 \\
            GT + $\mathcal{U}(50)$ & 96.2 & \textbf{50.9} & 51.0 & 51.0 & 0.1 \\
            GT + $\mathcal{U}(100)$ & 178.2 & \textbf{89.4} & 91.4 & 91.1 & 0.1 \\
            \bottomrule
        \end{tabular}
    }
    \caption{Denoising results on H3.6M dataset. We report MPJPE (mm) under Protocol \#2. ``Sep." here indicates model ``HPJ-T00". (Please refer to Section \ref{sec:supp_task}) ``Base" represents the MPJPE(mm) of noisy data. $\mathcal{N}$ and $\mathcal{U}$ denote Gaussian and uniform noise, respectively. $T$ denotes the start time of RSDE.}
    \label{table:supp_denoising}
\end{table}

\begin{table}
    \centering
    \small
    \setlength{\tabcolsep}{0.4cm}
        \begin{tabular}{l c c c}
            \toprule
            Model & Sep. & U2D & U3D \\
            \midrule
            MPJPE & \textbf{58.1} & 60.8 & 59.7 \\
            \bottomrule
        \end{tabular}
    \caption{Quality comparison of generated poses. We train 3 deterministic pose estimators on the poses generated by 3 different GFPose models (Sep. here means ``HPJ-T00". Please refer to Section \ref{sec:supp_task}). Then we evaluate them on the H3.6M test set. MPJPE(mm) under Protocol \#1 is reported.}
    \label{table:supp_generation}
\end{table}

\section{Different Sampling Approaches}
\label{sec:supp_imputation}
Imputation~\cite{song2020score} provides a flexible way to use score models for imputation tasks, like image inpainting and colorization. By repeatedly replacing part of the data in $\pose(t)$ with the observed data $\pose_{obs}$ during the sampling process, imputation allows us to use unconditional score models to impute the missing dimensions of data. Here, we can also view the 3D pose estimation and the completion task as imputation tasks, \ie, we impute the depth of 2D poses or impute the missing body joints. We compare the performance of imputation, conditional inference and the combination of these 2 techniques in Table~\ref{table:supp_impt_hpe3d}, \ref{table:supp_impt_complete_3d}, \ref{table:supp_impt_complete_2d}. We can get the following observations: (1) conditional inference generally performs better than imputation. (2) Combining two inference approaches can yield consistently better results than either approach alone on pose completion tasks. (3) Conditional inference performs best on the 3D human pose estimation task. (4) Imputation works better with unconditional models (HPJ-T00).

\section{More Qualitative Results}
\begin{figure*}[!t]
    \centering
    \includegraphics[width=0.97\linewidth]{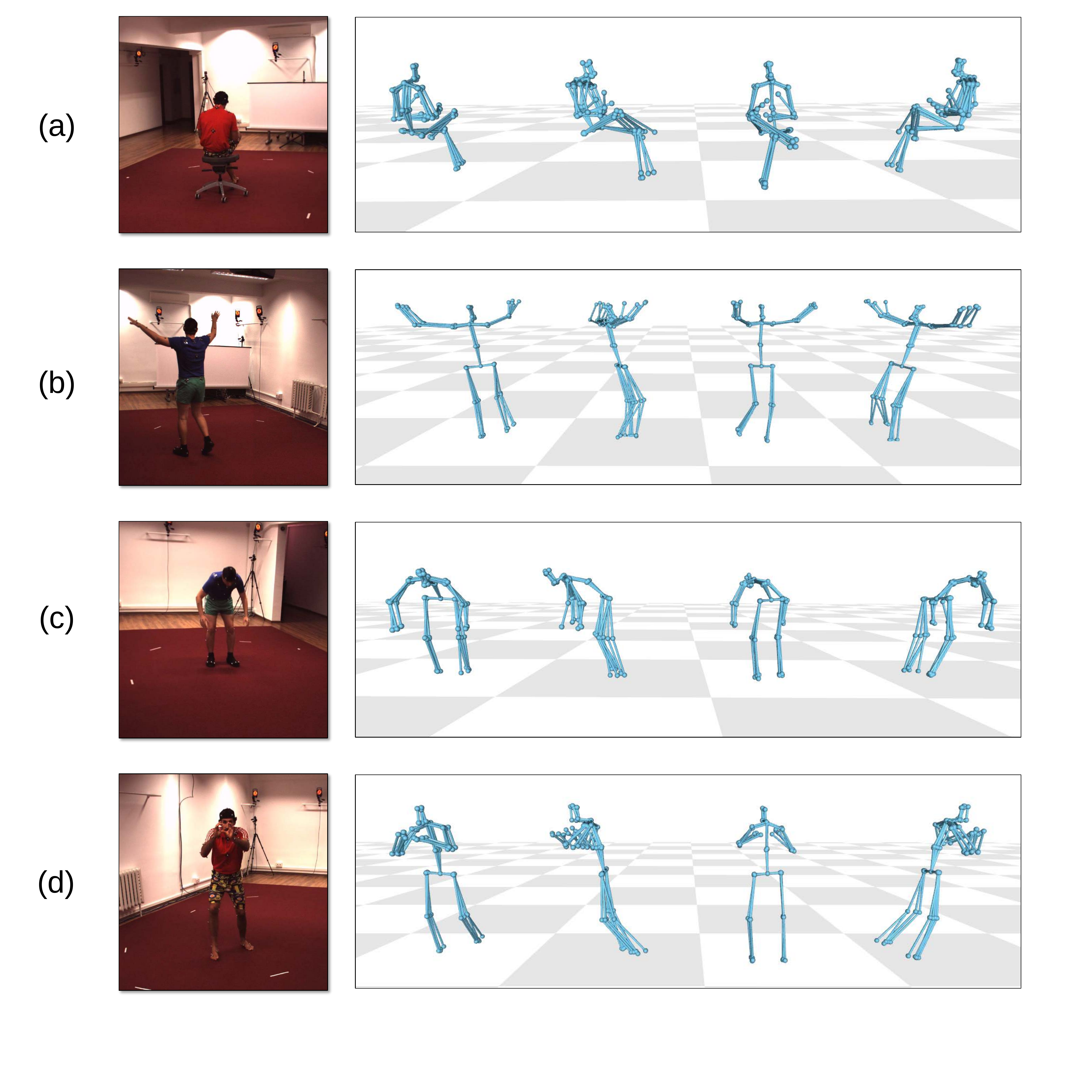}
    \caption{Multi-hypothesis 3D human pose estimation. We randomly sample $5$ hypotheses from GFPose and show them from $4$ different viewpoints. From left to right, the plausible poses are rotated 90 degrees clockwise around the z-axis.}
    \label{fig:supp_qua_estimation}
\end{figure*}

\begin{figure*}[t]
    \centering
    \includegraphics[width=0.97\linewidth]{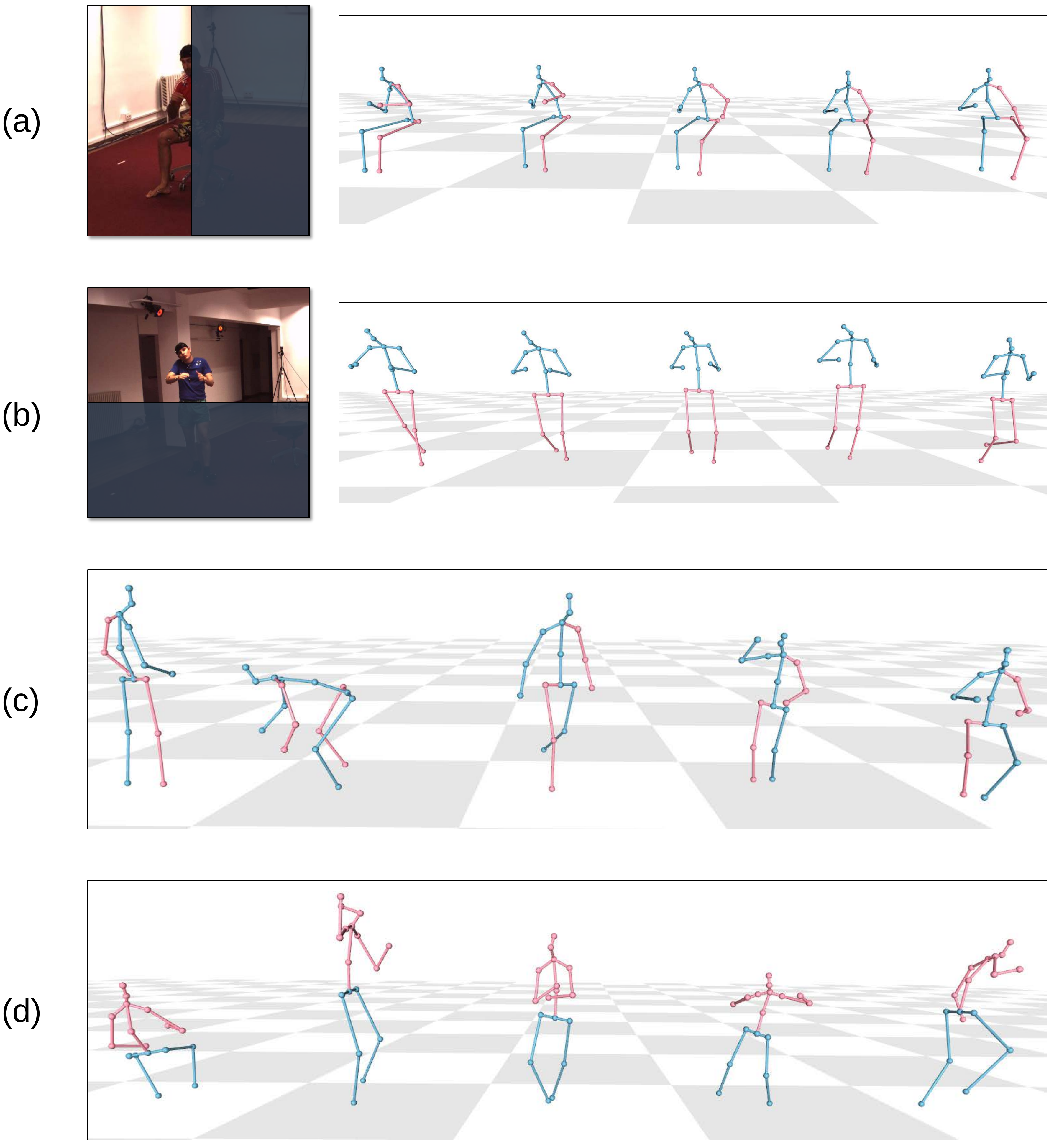}
    \caption{Pose completion. GFPose can recover full 3D human body from partial observations. 3D poses corresponding to the \textcolor[RGB]{97,165,194}{visible observation} and \textcolor[RGB]{255,143,163}{missing observation} are plotted in different colors for clarity. \textbf{(a)(b)} Recover full 3D human body from partial 2D image. We show five plausible poses sampled from GFPose. (a) \textcolor[RGB]{255,143,163}{Left side of the body} is invisible. (b) \textcolor[RGB]{255,143,163}{Lower body} is invisible. \textbf{(c)(d)} Recover full 3D human body from partial 3D poses. For each instance, we sample and plot one plausible completion. (c) \textcolor[RGB]{255,143,163}{Left arm and right leg} are missing. They are completed by GFPose. (d) \textcolor[RGB]{255,143,163}{Upper body} are missing and completed by GFPose. }
    \label{fig:supp_qua_completion}
\end{figure*}

\begin{figure*}[t]
    \centering
    \includegraphics[width=0.97\linewidth]{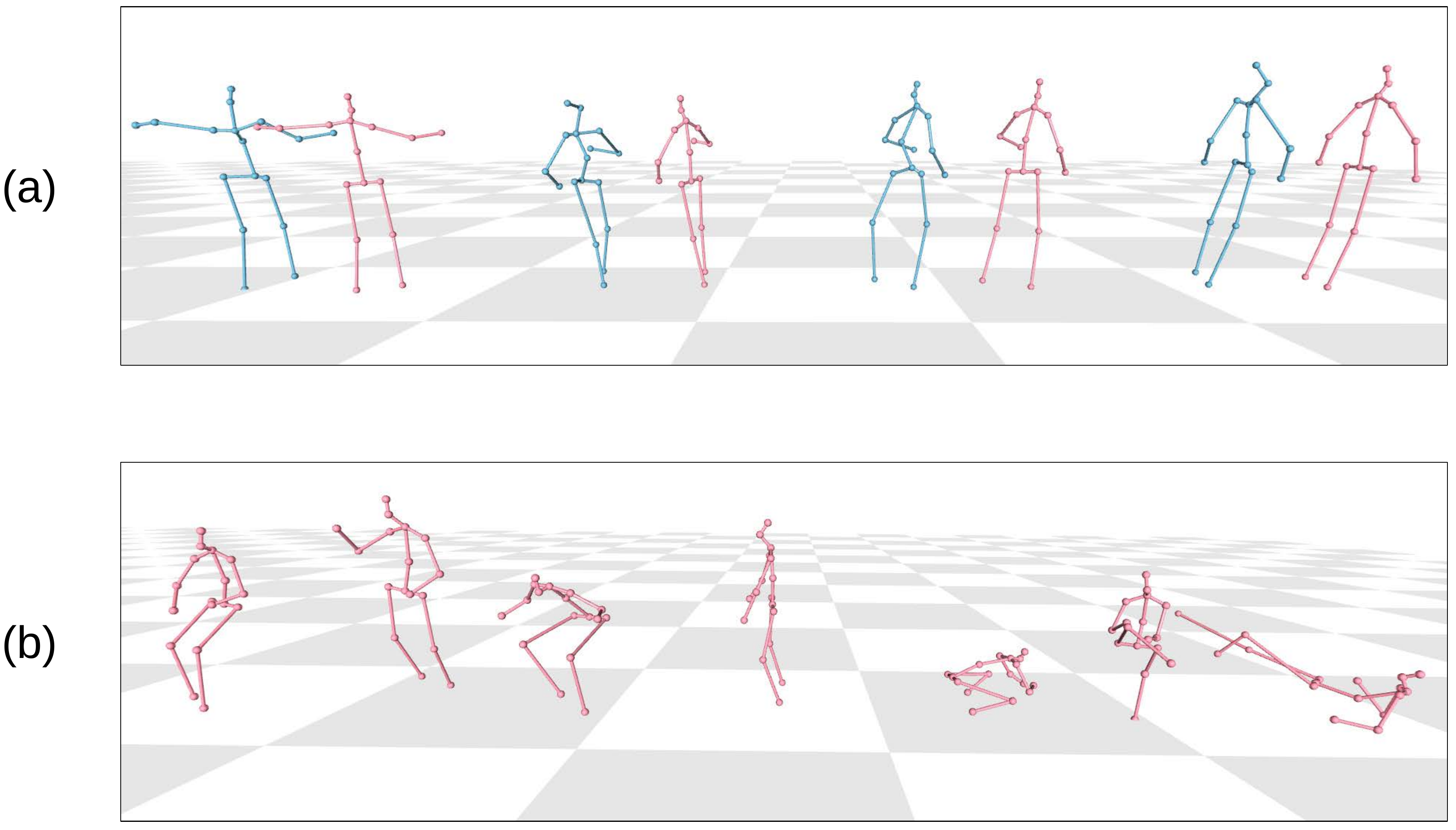}
    \caption{Pose denoising and generation. \textbf{(a)} Pose denoising. We add Gaussian noise $\sim \mathcal{N}(0, 100)$ onto GT and denoise it with GFPose. \textcolor[RGB]{97,165,194}{Nosiy poses} and \textcolor[RGB]{255,143,163}{denoised poses} are plotted in different colors. GFPose can effectively correct unreasonable poses. \textbf{(b)} Pose generation. GFPose can generate diverse and realistic poses. }
    \label{fig:supp_qua_rest}
\end{figure*}

We show more qualitative results for multi-hypothesis 3D pose estimation (Figure \ref{fig:supp_qua_estimation}), pose completion (Figure \ref{fig:supp_qua_completion}), pose denoising and generation (Figure \ref{fig:supp_qua_rest}). All poses are sampled from the unified  model ``U3D".



\begin{table}
    \centering
    \small
        \begin{tabular}{l c c c}
            \toprule
            Model & Cond. & Impt. & Cond.+Impt. \\
            \midrule
            HPJ-010 &  \textbf{35.1} & 49.5 & 36.8 \\
            HPJ-T00 &  - & \textbf{42.3} & - \\
            U2D &  \textbf{36.7} & 43.0 & 37.2 \\
            U3D &  \textbf{35.9} & 43.0 & 37.1\\
            \bottomrule
        \end{tabular}
    \caption{Comparison between different sampling approaches on 3D pose estimation. HPJ-T00 means we always activate the human level mask and learn an unconditional score model.}
    \label{table:supp_impt_hpe3d}
\end{table}

\begin{table}
    \centering
    \small
        \begin{tabular}{l c c c}
            \toprule
            Model & Cond. & Impt. & Cond.+Impt. \\
            \midrule
            HPJ-020 &  53.9 & 72.8 & \textbf{53.5} \\
            HPJ-T00 &  - & \textbf{64.9} & - \\
            U2D &  55.8 & 67.4 & \textbf{54.9} \\
            U3D & 54.6 & 66.8 & \textbf{53.8} \\
            \bottomrule
        \end{tabular}
    \caption{Comparison between different sampling approaches on pose completion from partial 2D observation (2 legs). HPJ-T00 indicates an unconditional score model.}
    \label{table:supp_impt_complete_2d}
\end{table}

\begin{table}
    \centering
    \small
        \begin{tabular}{l c c c}
            \toprule
            Model & Cond. & Impt. & Cond.+Impt. \\
            \midrule
            HPJ-020 &  5.3 & 8.3 & \textbf{5.2} \\
            HPJ-T00 &  - & \textbf{6.0} & - \\
            U2D &  - & \textbf{6.2} & - \\
            U3D &  8.6 & 6.1 & \textbf{5.6} \\
            \bottomrule
        \end{tabular}
    \caption{Comparison between different sampling approaches on pose completion from partial 3D observation (rigth leg). HPJ-T00 indicates an unconditional score model.}
    \label{table:supp_impt_complete_3d}
\end{table}

\end{document}